# DUAL-FISHEYE LENS STITCHING FOR 360-DEGREE IMAGING

*Tuan Ho* [1*]    *Madhukar Budagavi* [2]

[1]University of Texas – Arlington, Department of Electrical Engineering, Arlington, Texas, U.S.A
[2]Samsung Research America, Richardson, Texas, U.S.A

## ABSTRACT

Dual-fisheye lens cameras have been increasingly used for 360-degree immersive imaging. However, the limited overlapping field of views and misalignment between the two lenses give rise to visible discontinuities in the stitching boundaries. This paper introduces a novel method for dual-fisheye camera stitching that adaptively minimizes the discontinuities in the overlapping regions to generate full spherical 360-degree images. Results show that this approach can produce good quality stitched images for Samsung Gear 360 – a dual-fisheye camera, even with hard-to-stitch objects in the stitching borders.

## 1. INTRODUCTION

360-degree videos and images have become very popular with the advent of easy-to-use 360-degree viewers such as Cardboard [1] and GearVR [2]. This has led to renewed interest in convenient cameras for 360-degree capturing. A 360-degree image captures all the viewing directions simultaneously and gives users the sense of immersion when viewed. Early 360-degree imaging systems used a catadioptric camera [3], which combines lens (dioptric) and mirror (catoptric), to record 360-degree contents. Although the lens plus mirror geometry is sophisticated and usually requires proper calibration, such as one in [4][5], to generate good visual results, a catadioptric system can produce panoramas without seams. However, due to the inherent lens+mirror arrangement, the captured field of view is typically limited to less than 360x180 degrees, and some of the catadioptric systems are not compact.

An alternate method for 360-degree recording is using a polydioptric system which incorporates multiple wide-angle cameras with overlapping field of views. The images from the multiple cameras are stitched together to generate 360-degree pictures. However, due to camera parallax, stitching artifacts are typically observed at the stitching boundaries. Example 360-degree polydioptric cameras include Ozo [6], Odyssey [7], and Surround360 [8] by some of the major companies. The number of cameras used in these systems ranges from 8 to 17. These cameras typically deliver professional quality, high-resolution 360-degree videos.

---

*This research was done when the author interned at Samsung.

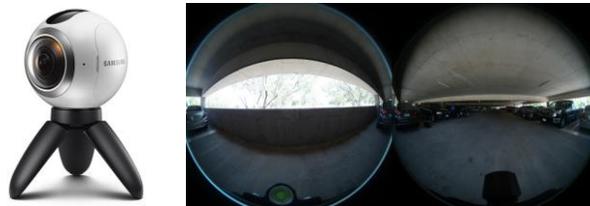

**Fig. 1.** (a) Samsung Gear 360 Camera (left). (b) Gear 360 dual–fisheye output image (7776 x 3888 pixels) (right). Left half: image taken by the front lens. Right half: image taken by the rear lens.

On the downside, these high-end 360-degree cameras are bulky and extremely expensive, even with the decreasing cost of image sensors, and are out of reach for most of the regular users. To bring the immersive photography experience to the masses, Samsung has presented Gear 360 camera, shown in **Fig. 1**(a). To make the camera compact, Gear 360 uses only two fisheye lenses whose field of view is close to 195 degrees each. The images generated by the two fisheye lenses (**Fig. 1**(b)) have very limited overlapping field of views but can, however, be stitched together to produce a full spherical 360x180 panorama.

For stitching of images from the multiple cameras, a feature-based stitching algorithm [9][10] is typically used to extract the features of the images being stitched. These features are then matched together. An iterative method is carried out to eliminate the incorrect matches (outliers). The reliability of this process not only depends on the iterative method being used but also the size of the overlapping regions. With sufficient overlap, more reliable matches (inliers) are retained while outliers get removed. Using these inliers, a homography matrix is computed to warp and register the pictures together (assuming the camera matrix is already estimated) before stitching them.

However, this conventional stitching method does not work well on Gear 360-produced pictures since there is very limited overlap between the two fisheye images. **Fig. 2** shows the stitching processes for the photos taken by the regular rectilinear lens and the ones taken by Samsung Gear 360. The pictures on the left column, from [11], in **Fig. 2** have a good overlap and can be aligned and stitched well. In contrast, Gear

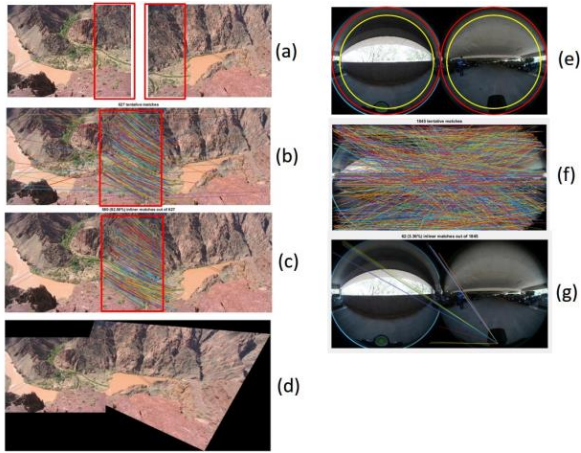

**Fig. 2.** Image stitching illustration. Left column: (a) Regular pictures with good overlaps. (b)(c) Features Matching using SIFT and outlier removal using RANSAC. (d) Image warping and panorama creation. Right column: (e) Fisheye images taken by Samsung Gear 360. (f)(g) Features Matching (using SIFT) and outlier removal (using RANSAC). Courtesy: VLFeat [11] toolbox

360 has limited overlap leading to a small number of inlier matches only on the outer ring of the fisheye images. This results in a homography matrix that is invalid for the interior of the fisheye images. Hence, a conventional stitching process cannot be directly used for stitching fisheye images from two-lens systems such as Gear 360.

This paper introduces a novel stitching method that adaptively minimizes the discontinuities in the overlapping regions of Gear 360 images to align and stitch them together. The proposed algorithm has four steps. The first step describes how to measure and compensate for the intensity fall off of the camera's fisheye lenses. The second phase explains the geometry transformation to unwarp the fisheye images to a spherical 2-Dimensional (equirectangular projection [12]) image. The next stage introduces our proposed two-step alignment to register the fisheye unwarped images. Finally, the aligned images are blended to create a full spherical 360x180-degree panorama.

## 2. DUAL–FISHEYE STITCHING

### 2.1. Fisheye Lens Intensity Compensation

Vignetting is an optical phenomenon in which the intensity of the image reduces at the periphery compared to the center. To compensate for this light fall-off, we captured an image of a large white blank paper using Gear 360 and measured the variation of pixel intensity along the radius of the fisheye image toward its periphery in **Fig. 3**. The intensity is normalized to one at the center of the picture. We used a polynomial function $p$ to fit the light fall-off data

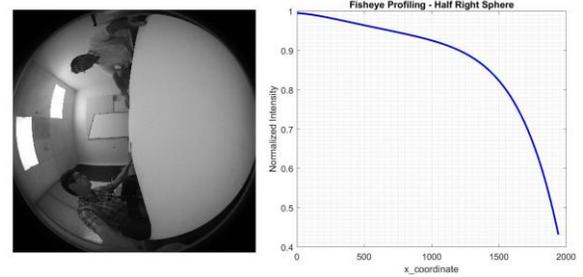

**Fig. 3.** (a) Fisheye profiling experiment (left). (b) Intensity fall-off curve (right)

$$p(x) = p_1 x^n + p_2 x^{n-1} + \cdots + p_n x + p_{n+1}$$

where $x$ is the radius from the center of the image.

### 2.2. Fisheye Unwarping

Fisheye lenses can produce ultra-wide field of views by bending the incident lights. As a result, the image looks severely distorted, particularly in the periphery. Therefore, a fisheye unwarping–a geometric transformation is necessary to generate a natural appearance for the Gear 360 fisheye-produced pictures. Instead of rectifying the fisheye-distorted image, we use a method that unwarps the image and returns a 2-D spherical projected picture for 360-degree purposes.

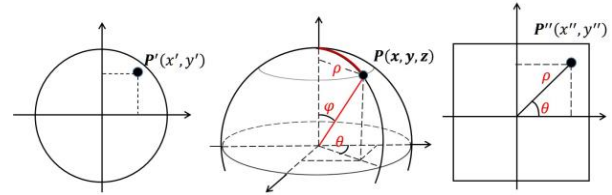

**Fig. 4.** Fisheye Unwarping

This method involves two steps as shown in **Fig. 4**. First, each point $P'(x', y')$ in the input fisheye image is projected to a 3-D point $P(\cos \varphi_s \sin \theta_s, \cos \varphi_s \cos \theta_s, \sin \varphi_s)$ in the unit sphere. $\varphi_s$ and $\theta_s$ can be derived by considering the coordinates of the fisheye image directly as pitch and yaw. Therefore, $\theta_s = f \frac{x'}{W} - 0.5$, and $\varphi_s = f \frac{y'}{H} - 0.5$, where $f$ is the lens' field of view (in degree), $W$ and $H$ are image width and height respectively. The second step derives the distance between the projected center and the 3-D point $P(x, y, z)$: $\rho = \frac{H}{f} \tan^{-1} \frac{\sqrt{x^2+z^2}}{y}$, $x = \cos \varphi_s \sin \theta_s$, $y = \cos \varphi_s \cos \theta_s$, $z = \sin \varphi_s$. Then the 2-D spherical (equirectangular) projected point $P''(x'', y'')$ is constructed as $x'' = 0.5W + \rho \cos \theta$, $y'' = 0.5H + \rho \sin \theta$, and $\theta = \tan^{-1}(z/x)$. In this equirectangular projection, $x''$ and $y''$ are pitch and yaw respectively. The unwarped image can be viewed on a 360-degree player.

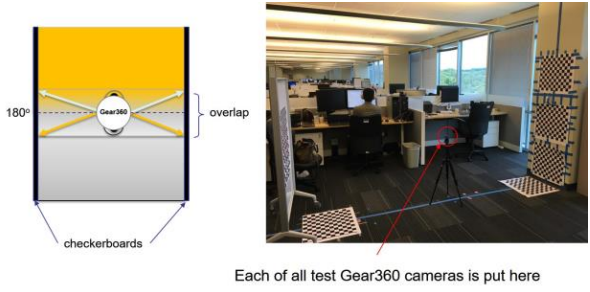

**Fig. 5.** Experiment Setup for Gear 360 Dual-lens Misalignment

### 2.3. Fisheye Lens Distortion Measurement

To measure the fisheye lens distortion, we used Gear 360 camera to take multiple photos of checkerboard patterns and adopted a calibration process in [4][5] to calibrate the lens. The affine matrix found by this process indicates that the Gear 360 fisheye lens distortion is negligible. Thus, no correction is needed.

### 2.4. Blending

This paper employs a ramp function to blend the overlapping regions between two unwarped images. The blended pixel $B(r, c)$ at row $r$ and column $c$ in the overlapping region of size $r \times n$ is computed as:

$$B(r, c) = \alpha_1 * L(r, c) + \alpha_2 * R(r, c)$$

Where $L(r, c)$ and $R(r, c)$ are the pixels at position $(r, c)$ taken from the left image and the right image respectively.

When blending the right overlapping area, $\alpha_1 = c/n$, $\alpha_2 = (n - c + 1)/n$. $n$ is the width of the blending region.

When blending the left overlapping area, $\alpha_1 = (n - c + 1)/n$, $\alpha_2 = c/n$.

### 2.5. Two-step Alignment

#### 2.5.1. Lens Misalignment Compensation

After unwarping, the two images are not aligned with each other. The between-lenses misalignment patterns are similar for different Gear 360 cameras of the same model. To minimize this geometric misalignment we propose to use a control-point-based approach as follows.

In the setup in **Fig. 5**, we position the Gear 360 so that both the lenses see the checkerboards on their sides. Therefore, they have the same view of the overlapping regions. Also, the distance between the camera and the checkerboards is around 2m, which is about the maximum reach that the checkerboard corners are still clearly visible for control point selection. The images taken by the Gear 360 left and right lenses are unwarped using the method in section 2.2, and arranged in 360x180-degree planes in **Fig. 6**. About 200 pairs of control points are then manually selected from the overlapping regions between the unwarped pictures, and are used to estimate an affine matrix $A$, which warps a point $B(x_2, y_2)$ to $T(x_1, y_1)$ as follows:

$$[x_1 \; y_1 \; 1] = [x_2 \; y_2 \; 1] \, A, \text{where } A = \begin{bmatrix} a & b & 0 \\ c & d & 0 \\ t_x & t_y & 1 \end{bmatrix}$$

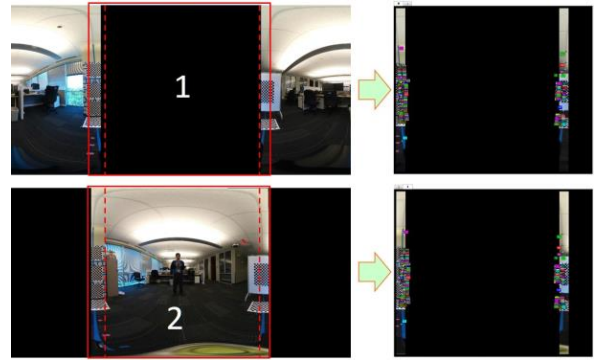

**Fig. 6.** Control point selection on the overlapping area. The fisheye images are unwarped using the method presented in section 2.2

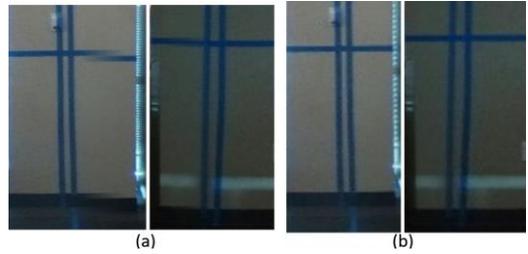

**Fig. 7.** (a) Without the first alignment. (b) With the first alignment

#### 2.5.2. Refined Alignment

The first registration helps align the images as shown in **Fig. 7**. However, when the objects in the boundaries move closer or further away from the camera, the horizontal discontinuities become visible as shown in **Fig. 8**(c).

To minimize the discontinuity in the overlapping regions, we choose to maximize the similarity in these areas. To this end, this paper proposes a novel adaptive alignment that involves a fast template matching for objects in the overlapping region and utilizes the matching displacement to derive a refined affine matrix to align the images further.

The matching is a normalized cross-correlation operation. The cross-correlation of two signals maximizes at a point when the two signals match each other. In addition, since there is always some level of exposure differences in the overlapping regions, the template and reference images to be matched should be normalized. Therefore, this proposal employs a fast normalized cross-correlation algorithm in [13]:

$$\gamma(u, v) = \frac{\sum_{x,y}[f(x, y) - \overline{f_{u,v}}][t(x - u, y - v) - \bar{t}]}{\left\{\sum_{x,y}[f(x, y) - \overline{f_{u,v}}]^2 \sum_{x,y}[t(x - u, y - v) - \bar{t}]^2\right\}^{0.5}}$$

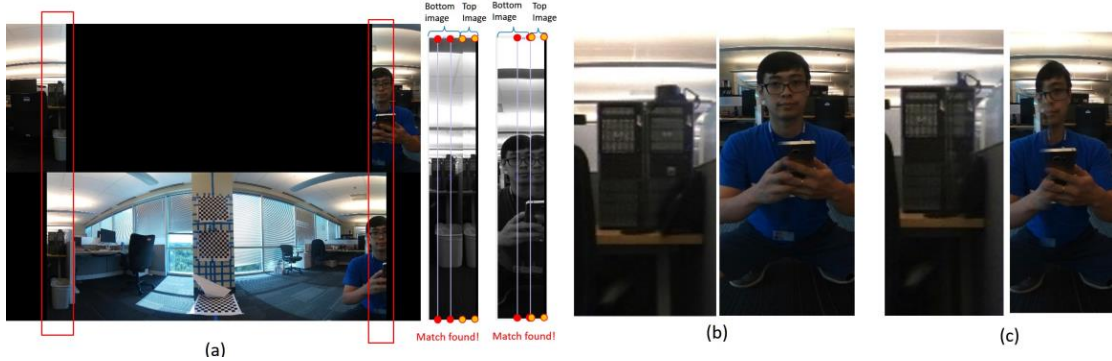

**Fig. 8.** (a) A person close to the camera and between the lens boundary. (b) The blended overlaps with the proposed refined alignment (discontinuity minimized). (c) The blended overlaps without the proposed refined alignment (very visible discontinuity). The first alignment is already applied for both (b) and (c) to align the images vertically.

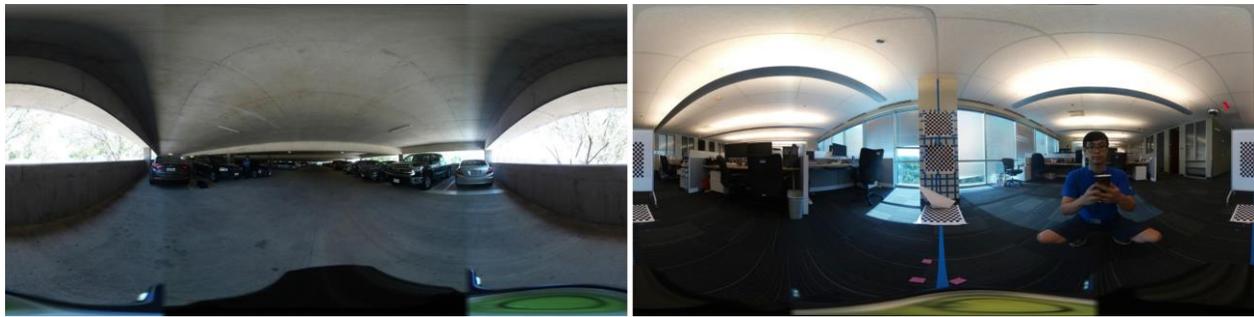

**Fig. 9**. Samsung Gear 360's 360x180-degree panorama stitched by this proposed method. Objects are far away (left) and very close to camera (right)

where $\gamma$ is the normalized cross-correlation, $f$ is the reference image, $\bar{t}$ is mean of the template image, $\overline{f_{u,v}}$ is the mean of $f(x,y)$ in the region under the template. The template and reference are taken from the top and bottom unwarped images respectively, as shown in **Fig. 8** (a).

The maximum value of the normalized cross-correlation returns the displacement of where the best match occurs. This shift indicates how much the template – a rectangular window should move to match the reference. The proposed method then estimates an affine matrix from vertices of the matching windows (four in each overlapping region) and warp the bottom image to align it further with the top one.

**Fig. 8** shows that the refined alignment helps align the images by maximizing the similarity in the overlapping region. The person, close and in the lens boundary, appears as a complete one (i.e. no visible duplicate or missing any body parts) in the stitched 360x180-degree picture.

### 3. IMPLEMENTATION & RESULTS

We have implemented the proposed approach in C++ with OpenCV library and Matlab. The affine matrix in the first alignment is precomputed and included as part of the fisheye unwarping process. The refined alignment, however, is computed on-line, adaptively to the scene. The polynomial coefficients in section 2.1 are: $p_1 = -7.5625\times10^{-17}$, $p_2 = 1.9589\times10^{-13}$, $p_3 = -1.8547\times10^{-10}$, $p_4 = 6.1997\times10^{-8}$, $p_5 = -6.9432\times10^{-5}$, $p_6 = 0.9976$. We found that the field of view of 193 degrees, which is very close to the documented 195-degree, gives the best results as shown in **Fig. 9**. Our approach can also accurately stitch images taken by different Gear 360 cameras of same model thanks to the proposed refined alignment that operates adaptively and can compensate for the geometric mismatch between Gear 360 lenses.

### 4. CONCLUSION

This paper introduces a new method for stitching images from 360-degree cameras with dual-fisheye lens. It uses a novel alignment algorithm that adaptively maximizes the similarities in the boundary regions of the images from the two fisheye lenses for accurate registration and stitching. In summary, the proposed approach compensates for fisheye lens' intensity fall-off, unwarps the fisheye images, then registers them together using the proposed adaptive alignment, and applies blending on the registered images to create a 360x180-degree panorama that is viewable on 360-degree players. Results show that not only this method can stitch Gear 360 images that have limited overlap, but it can also produce well-stitched pictures even if there are objects that are at an arbitrary distance to the camera and stand in the lenses boundaries.